# Forecasting the Future with Future Technologies:
## Advancements in Large Meteorological Models

Hailong Shu, Yue Wang*, Weiwei Song*, Huichuang Guo, Zhen Song
State Key Laboratory of NBC Protection for Civilian
Beijing, China
* Corresponding author: wy.1998.8@163.com; songweiwei_1981@126.com

*Abstract*—The field of meteorological forecasting has undergone a significant transformation with the integration of large models, especially those employing deep learning techniques. This paper reviews the advancements and applications of these models in weather prediction, emphasizing their role in transforming traditional forecasting methods. Models like FourCastNet, Pangu-Weather, GraphCast, ClimaX, and FengWu have made notable contributions by providing accurate, high-resolution forecasts, surpassing the capabilities of traditional Numerical Weather Prediction (NWP) models. These models utilize advanced neural network architectures, such as Convolutional Neural Networks (CNNs), Graph Neural Networks (GNNs), and Transformers, to process diverse meteorological data, enhancing predictive accuracy across various time scales and spatial resolutions. The paper addresses challenges in this domain, including data acquisition and computational demands, and explores future opportunities for model optimization and hardware advancements. It underscores the integration of artificial intelligence with conventional meteorological techniques, promising improved weather prediction accuracy and a significant contribution to addressing climate-related challenges. This synergy positions large models as pivotal in the evolving landscape of meteorological forecasting.

## I. INTRODUCTION

Meteorological forecasting, a cornerstone of modern society, profoundly influences diverse sectors, including agriculture, transportation, disaster management, and public safety. The capacity for precise weather prediction is not merely a matter of convenience but a critical tool for effective planning and decision-making. It plays a vital role in mitigating the detrimental impacts of extreme weather phenomena and optimizing conditions for various human activities. Moreover, it contributes significantly to our long-term understanding of climate change and its multifaceted implications [1].

Historically, the domain of meteorological forecasting has been underpinned by numerical weather prediction (NWP) models. These models, grounded in mathematical representations of atmospheric processes, have been instrumental in enhancing our predictive capabilities. However, despite their advancements, traditional NWP models are not without limitations. They often necessitate substantial computational resources, which hampers their ability to deliver high-resolution forecasts promptly. Additionally, these models face challenges in accurately capturing complex atmospheric dynamics and microscale meteorological events, primarily due to the inherent simplifications and parameterizations within their mathematical frameworks [2].

The advent of large-scale, data-driven models marks a significant evolution in the field of meteorological forecasting. These models, powered by the advancements in artificial intelligence (AI) and, more specifically, deep learning, harness extensive and varied datasets to discern and forecast atmospheric behavior. In contrast to traditional models that predominantly depend on physical equations, AI-based models derive insights directly from historical and real-time meteorological data. This approach holds the promise of enhanced accuracy in predictions, particularly for small-scale weather phenomena. The integration of these large models into meteorological forecasting heralds new possibilities for improving forecast precision, diminishing computational demands, and offering more detailed insights into weather patterns [3].

This paradigm shift in meteorological forecasting, facilitated by large models, signifies a transformative change in our approach to understanding and predicting weather. It suggests not only an augmentation of existing capabilities but also, in certain aspects, a potential to surpass the traditional NWP models [4].

The structure of the remainder of this paper is as follows: Section 2 delves into the evolution and current state of large models in meteorological forecasting. Section 3 examines specific deep learning architectures employed in these models, including Transformers and Graph Neural Networks. Section 4 addresses the challenges and limitations inherent in these models, while also outlining prospective avenues for future research. Finally, Section 5 concludes by summarizing the key findings and underscoring the potential impact of large models on the landscape of meteorological forecasting.

## II. BASIC CONCEPTS AND TECHNOLOGIES OF METEOROLOGICAL LARGE MODELS

Central to the advancements in meteorological large models are the concepts of deep learning and neural networks, key subsets of artificial intelligence (AI) [5]. Deep learning, characterized by its use of multi-layered neural networks, enables these models to autonomously learn and make intelligent decisions. The networks, comprising various nodes and layers, are designed to emulate the human brain's structure and function, facilitating the processing of large datasets and recognition of intricate patterns in meteorological data.

The architecture of these models plays a pivotal role in their capability to process and interpret data. Convolutional Neural Networks (CNNs), known for their proficiency in handling spatial data, are extensively utilized in meteorological forecasting. Their ability to effectively analyze weather patterns across diverse geographic regions makes them an invaluable tool in this field. Graph Neural Networks (GNNs) represent another significant architectural approach. They excel in managing data structured in graph formats, proving particularly useful for modeling the complex interconnections within atmospheric variables [6].

Additionally, the Transformer architecture, renowned for its efficiency in sequential data processing, has been increasingly adopted in meteorological forecasting. Its capacity to capture and interpret the temporal dynamics of weather data further enhances the predictive accuracy of these models [7]. Each of these architectures brings unique strengths to the table, and their deployment is tailored to align with the specific demands and challenges of various meteorological forecasting tasks.

In sum, the integration of these sophisticated AI technologies into meteorological large models marks a significant leap forward in our ability to forecast weather patterns with greater precision and detail. The continuous evolution and optimization of these technologies promise further enhancements in the accuracy and efficiency of weather prediction.

### III. COMPREHENSIVE REVIEW OF LARGE METEOROLOGICAL MODELS

#### A. Overview of Pioneering Large Meteorological Models

*1) FourCastNet[8]:*

FourCastNet signifies a revolutionary step in meteorological forecasting, integrating cutting-edge techniques such as the Fourier Neural Operator (FNO) and Vision Transformer (ViT). This combination, adept at handling high-resolution inputs, empowers FourCastNet to proficiently predict fast-timescale variables like surface wind speed, precipitation, and atmospheric water vapor. Utilizing the extensive ERA5 dataset, which provides hourly estimates dating from 1979 to the present, FourCastNet has a solid foundation for both training and validation, ensuring robust predictive capabilities.

FourCastNet is specifically optimized for short to medium-range forecasting, showcasing exceptional skill in predicting small-scale variables, particularly precipitation. This advanced architecture allows it to achieve high levels of accuracy, often surpassing traditional models like the ECMWF Integrated Forecasting System (IFS) in forecasting large-scale variables. This model's precision is crucial in meteorological forecasting, a field where accuracy is imperative. Additionally, FourCastNet's remarkable speed in generating forecasts, capable of producing week-long forecasts in under two seconds, underlines its potential to revolutionize operational meteorology with quick, reliable predictions aiding critical weather-related decision-making.

The integration of advanced neural network technologies in FourCastNet marks a significant milestone in AI-driven weather forecasting. Its innovative approach not only sets new standards in the realm of meteorological prediction but also offers a glimpse into the future possibilities of weather forecasting, where AI's role is becoming increasingly prominent.

*2) Pangu-Weather[9]:*

Pangu-Weather represents a significant advancement in AI-driven meteorological modeling, utilizing the innovative 3D Earth Specific Transformer (3DEST) architecture. With its training grounded in the ERA5 dataset, which encompasses global weather data from 1979 to 2017, the model offers an extensive understanding of weather patterns. This particular architecture is tailored to interpret complex atmospheric data effectively, distinguishing Pangu-Weather in the field of meteorological forecasting.

This model's range of capabilities covers a wide spectrum of atmospheric and surface weather variables, including geopotential, specific humidity, temperature, and wind speed. Its ability to forecast from an hour to a week suits a variety of forecasting needs, from immediate to extended predictions. Pangu-Weather's groundbreaking approach has successfully surpassed conventional numerical weather prediction methods in accuracy, marking a pivotal moment in the evolution of weather forecasting.

Moreover, Pangu-Weather's role in predicting extreme weather events and aiding large-member ensemble forecasts in real-time demonstrates its practical utility. The model's robustness and adaptability in diverse weather scenarios promise significant enhancements in both short and medium-range weather forecasting. This advancement solidifies Pangu-Weather as a valuable tool in the ongoing quest to refine meteorological prediction accuracy and reliability.

*3) FengWu[10]:*

FengWu introduces a novel approach to medium-range weather forecasting, employing a transformer-based architecture with modal-customized encoder-decoders. Trained on the ERA5 dataset, which includes 6-hourly sampled data from 1979 to 2015, FengWu is equipped with a substantial foundation for understanding and predicting atmospheric dynamics. The model excels in forecasting a comprehensive range of variables across 37 vertical levels, covering both atmospheric and surface conditions.

The unique design of FengWu focuses on multi-modal and multi-task processing, utilizing a region-adaptive uncertainty loss to balance the optimization of various predictors. This leads to significant improvements in forecasting accuracy, particularly for medium-range forecasts. The model's success in surpassing other models like GraphCast in several key metrics, such as reducing root mean square error in crucial predictions like the 10-day lead global z500 forecast, exemplifies its capability as a significant advancement in the field.

In addition to its technical achievements, FengWu's practical implications in meteorological forecasting are noteworthy. Its ability to extend skillful global medium-range weather forecasts beyond traditional limits showcases its potential as a game-changer in the field. The model's efficiency in generating forecasts, coupled with its high accuracy, makes it an invaluable tool for meteorologists and weather prediction systems worldwide. FengWu's innovative approach and breakthroughs in medium-range forecasting highlight the evolving landscape of meteorological prediction, where AI and deep learning are playing increasingly vital roles.

*4) FuXi[11]:*

FuXi represents a breakthrough in meteorological forecasting, offering a 15-day global forecast system. This machine learning (ML) based system is structured as a cascade model, tailored to address the challenge of reducing forecast error accumulation over longer forecast periods. FuXi's design allows it to deliver forecasts that match, and in some cases exceed, the skillful forecast lead time of the European Centre for Medium-Range Weather Forecasts (ECMWF)'s high-resolution forecast (HRES). Notably, it extends the lead time for critical weather variables such as Z500 from 9.25 to 10.5 days and for T2M from 10 to 14.5 days.

Developed using 39 years of ECMWF ERA5 reanalysis data, FuXi operates with a spatial resolution of 0.25 ° and a temporal resolution of 6 hours. The ERA5 dataset, known for its comprehensive coverage and accuracy, includes hourly data of surface and upper-air parameters. FuXi focuses on predicting a total of 70 variables, which encompass 5 upper-air atmospheric variables at 13 pressure levels and 5 surface variables.

The architecture of the FuXi model includes three main components: cube embedding, U-Transformer, and a fully connected layer. The model handles both upper-air and surface variables, creating a data cube to effectively manage the high-dimensional input data. This cube embedding process significantly reduces the temporal and spatial dimensions of the input data, streamlining the forecasting process. The U-Transformer then processes the embedded data, and a simple fully connected layer is used for prediction. The outputs are fine-tuned for specific forecast periods and then combined in a cascaded manner, ensuring optimal performance across different time windows, ranging from short (0-5 days) to long (10-15 days) forecasts.

*5) ClimaX[12]:*

ClimaX stands as a transformative model in weather and climate science, powered by a transformer-based architecture and pre-training on CMIP6 data. This approach equips ClimaX to handle a diverse range of temporal forecasting tasks, adapting to both historical and real-time data inputs. Its versatility is showcased in its ability to provide predictions for tasks ranging from nowcasting to short and medium-range forecasting, making it a comprehensive tool for weather prediction.

The model's unique strength lies in its application across multiple scales, where it can predict key weather variables like geopotential at 500hPa, temperature at various altitudes, and zonal wind speed. This adaptability allows ClimaX to cater to a broad spectrum of forecasting needs, demonstrating superior performance in climate projections and weather forecasting benchmarks. The model's ability to perform effectively even when pre-trained at lower resolutions and compute budgets highlights its efficiency and the innovative use of deep learning in meteorological forecasting.

Furthermore, ClimaX's application in multi-scale meteorological forecasting signifies a major leap forward in the field. Its performance in different weather prediction tasks, ranging from immediate weather conditions to more extended forecasts, illustrates its potential as a versatile and robust tool.

The integration of advanced transformer-based technology in ClimaX sets a new standard in the domain of AI-driven weather forecasting, showcasing the significant impact deep learning can have in enhancing our understanding and prediction of climate and weather dynamics.

*6) GraphCast[13]:*

GraphCast emerges as a leading-edge model in meteorological forecasting, leveraging the power of Graph Neural Networks (GNNs) to decode and predict complex weather dynamics. Its utilization of a substantial 39-year dataset from the ECMWF's ERA5 reanalysis archive positions GraphCast to deliver detailed and precise medium-range forecasts. The model's GNN architecture is adept at analyzing intricate weather patterns, enabling it to offer forecasts with a horizontal resolution of 0.25 °across 13 vertical levels.

A standout feature of GraphCast is its swift processing speed, capable of generating accurate 10-day forecasts in less than a minute. This efficiency is crucial in operational meteorology, where time-sensitive and reliable predictions are essential. The model's proficiency in handling and swiftly processing large-scale data showcases the strides AI technologies have made in weather forecasting, offering substantial improvements over traditional forecasting methods.

GraphCast is not only a technological breakthrough but also a practical asset in global weather prediction. Its superiority in forecasting skill, particularly for medium-range predictions, enhances its value in the meteorological community. The model's ability to manage vast amounts of data and quickly adapt to new information positions it at the forefront of the field, highlighting the vast potential of integrating advanced AI technologies like GNNs into meteorological forecasting.

*B. Comparison of key indicators*

*1) Model Architecture*

These advanced meteorological forecasting models exhibit a diverse range of architectures, each tailored to specific forecasting challenges. FourCastNet integrates the Fourier Neural Operator (FNO) with a Vision Transformer (ViT) backbone, adept at handling high-resolution inputs for predicting fast-timescale variables like surface wind speed and precipitation. Pangu-Weather employs the innovative 3D Earth Specific Transformer (3DEST), designed to effectively interpret complex atmospheric data. GraphCast leverages Graph Neural Networks (GNNs) to analyze intricate weather patterns, offering detailed medium-range forecasts. ClimaX, with its transformer-based architecture, excels in a variety of forecasting tasks, from nowcasting to medium-range predictions. FengWu utilizes a transformer-based architecture with modal-customized encoder-decoders, focusing on multi-modal and multi-task processing. FuXi, a cascaded machine learning system, is structured to reduce forecast error accumulation over extended periods. Each model's architecture reflects its unique approach to addressing the complexities of meteorological forecasting, showcasing the potential of AI and deep learning in enhancing predictive accuracy and efficiency in this field.

*2) Training Cost*

These cutting-edge weather forecasting models demonstrate significant differences in training costs, particularly in terms of GPU requirements and computational time. FourCastNet's training, conducted on 64 Nvidia A100 GPUs for approximately 16 hours, indicates a substantial demand for high-performance computing resources while also reflecting relatively high computational efficiency. Pangu-Weather, trained on 192 NVIDIA Tesla-V100 GPUs for about 15 days, not only reveals its need for a large computational resource but also suggests the computational intensity of its complex 3D Earth Specific Transformer architecture in processing extensive data. FengWu's training, carried out on 32 Nvidia A100 GPUs for 17 days, indicates high computational demands for handling multi-modal and multi-task processes and a lengthy training period. FuXi, with a lighter training load on 8 Nvidia A100 GPUs for around 30 hours, may owe its efficiency to an optimized cascading model structure and effective training strategies. GraphCast underwent training on 32 Cloud TPU v4 devices for approximately four weeks, reflecting the high computational demands of its Graph Neural Network architecture in processing complex weather patterns over an extended period. The training costs of these models highlight the significant computational resources and time investment required for applying deep learning in meteorological forecasting, while also revealing the impact of different architectures and optimization strategies on training efficiency.

### 3) Specialization and Performance

Each model exhibits unique strengths in specialization and performance. FourCastNet excels in short to medium-range forecasting, particularly adept at predicting small-scale variables such as precipitation. Pangu-Weather, capable of forecasting a wide range of atmospheric and surface weather variables, focuses on accuracy in medium-range predictions. GraphCast specializes in medium-range forecasting with high accuracy and rapid processing capabilities. ClimaX demonstrates exceptional performance across various tasks, from nowcasting to short and medium-range forecasting. FengWu, focusing on medium-range weather forecasting, surpasses models like GraphCast in key metrics. FuXi offers a 15-day global forecast system, extending the predictive lead time for critical weather variables, indicating its proficiency in long-range forecasting.

### 4) Potential for Development and Future Applications

These models showcase immense diversity in their potential for development and future applications. FourCastNet's speed and accuracy herald its revolutionary potential in operational meteorology. Pangu-Weather's ability to surpass traditional methods marks a pivotal moment in the evolution of weather forecasting. GraphCast's efficiency and proficiency in handling large-scale data demonstrate the strides made by AI technologies in weather prediction. ClimaX's application in multi-scale meteorological forecasting signifies a major leap forward, with its multi-task adaptability indicating significant developmental potential. FengWu's innovative approach and breakthroughs in medium-range forecasting emphasize the evolving landscape of meteorological prediction. FuXi's design, reducing error accumulation in long-term forecasts, and its extended predictive lead time for key weather variables showcase its immense potential in long-range forecasting. These models, with their unique capabilities and advancements, highlight the growing role and impact of AI and deep learning in the realm of meteorological forecasting, paving the way for future innovations and applications in this field.

## IV. CHALLENGES AND OPPORTUNITIES OF METEOROLOGICAL LARGE MODELS

The integration of large models in meteorological forecasting brings with it several challenges. One primary challenge is data acquisition: obtaining comprehensive, high-quality, and real-time weather data is crucial for accurate forecasting, yet it is often a complex and resource-intensive process. Additionally, the sheer volume and variety of data necessitate advanced processing capabilities, which can be a bottleneck for many systems. Real-time performance is another critical issue; the ability to process data and provide forecasts quickly is essential, especially for extreme weather events, but this demands significant computational resources.

Despite these challenges, there are numerous opportunities for advancement in this field. One key area is model optimization. Enhancements in algorithms and learning techniques can lead to more efficient and accurate models that are better at capturing complex weather dynamics. Hardware development also presents significant opportunities. Advances in computing power, such as more powerful GPUs and specialized processors like TPUs, can dramatically improve the processing capabilities necessary for large models. Moreover, the integration of AI with traditional meteorological methods offers a promising avenue for creating more robust and accurate forecasting systems.

The future of large models in meteorological forecasting is poised at an exciting juncture. With ongoing advancements in AI, data processing, and computational hardware, these models are set to become even more integral to our ability to predict and understand weather patterns, thereby enhancing our capacity to respond to climate-related challenges.

## V. Conclusions

This paper has highlighted the significant advancements and burgeoning potential of large models in the field of meteorological forecasting. Utilizing deep learning techniques, such as Convolutional Neural Networks (CNNs), Graph Neural Networks (GNNs), and Transformers, these models have proven their efficacy in providing accurate, high-resolution forecasts across diverse ranges. The developments seen in models like FourCastNet, Pangu-Weather, GraphCast, ClimaX, and FengWu exemplify the considerable progress made in this domain, demonstrating a marked proficiency in navigating complex weather dynamics and predicting various meteorological phenomena with enhanced accuracy and efficiency.

Looking to the future, the potential of these models in meteorological forecasting appears boundless. With the continuous evolution of data acquisition methods and the expansion of computational capabilities, we can anticipate that these large models will offer increasingly refined and timely forecasts. Such advancements are poised not only to deepen our understanding of weather patterns but also to strengthen our preparedness for facing climate-related challenges.

A critical aspect of this evolution is the integration of artificial intelligence with traditional meteorological methods, forming a synergistic union. This integration combines the powerful data-processing capabilities of AI with the in-depth atmospheric science insights inherent in traditional methods. Adopting this collaborative approach promises to yield more robust and comprehensive forecasting systems. Consequently, this fusion is anticipated to significantly enhance our ability to predict and effectively respond to the dynamic and complex nature of global weather and climate phenomena.